\documentclass{article}

\usepackage{arxiv}

\usepackage[utf8]{inputenc}
\usepackage[T1]{fontenc}

\usepackage{microtype}
\usepackage{booktabs}
\usepackage{amsfonts}
\usepackage{nicefrac}

\usepackage{graphicx}
\usepackage{dblfloatfix}   
\usepackage{placeins}      
\usepackage{hyperref}
\usepackage{url}
\usepackage{cuted}     
\usepackage{caption}   
\usepackage{amsmath,amssymb}

\usepackage[numbers,sort&compress]{natbib}

\hypersetup{
  colorlinks=true,
  citecolor=blue,
  linkcolor=blue,
  urlcolor=blue
}

\newcommand{\DOI}[1]{\href{https://doi.org/#1}{doi:#1}}

\title{Region of Interest Segmentation \\ and Morphological Analysis
for Membranes in Cryo-Electron Tomography}

\renewcommand{\shorttitle}{TomoROIS-SurfORA: ROI Segmentation and Membrane Morphometry}

\author{
Xingyi Cheng$^{1,\text{\S}}$,
Julien Maufront$^{1}$,
Aur\'elie Di Cicco$^{1}$,
Dani\"el M.~Pelt$^{2}$,
Manuela Dezi$^{1,*}$,
Daniel L\'evy$^{1,*}$\\
\small
$^1$ Institut Curie, CNRS UMR168, Universit\'e PSL, Sorbonne Universit\'e, Paris, France\\
$^2$ Leiden Institute of Advanced Computer Science (LIACS), Universiteit Leiden, the Netherlands\\
\texttt{daniel.levy@curie.fr, manuela.dezi@curie.fr}\\
$\S$ Current address: Inria Center at University of Rennes, SAIRPICO team, Rennes, France
}

\date{February 20, 2026}

\begin{document}



\makeatletter
\twocolumn[
\@twocolumnfalse
\maketitle
\vspace{-0.8em}

\begin{abstract}
Cryo-electron tomography (cryo-ET) enables high-resolution, three-dimensional reconstruction of biological structures, including membranes and membrane proteins. Identification of regions of interest (ROIs) is central to scientific imaging, as it enables isolation and quantitative analysis of specific structural features within complex datasets. In practice, however, ROIs are typically derived indirectly through full structure segmentation followed by \emph{post hoc} analysis. This limitation is especially apparent for continuous and geometrically complex structures such as membranes, which are segmented as single entities. For example, identifying membrane contact sites (MCS) requires segmentation of all membranes, followed by manual inspection and curation before distance-based computations can be performed. In contrast, membrane invaginations lack a clear mathematical definition and must therefore be detected through learned representations. 

Here, we developed \emph{TomoROIS-SurfORA}, a two-step framework for direct, shape-agnostic ROI segmentation and morphological surface analysis. \emph{TomoROIS} performs deep learning-based ROI segmentation and can be trained from scratch using small annotated datasets, enabling practical application across diverse imaging data. \emph{SurfORA} processes segmented structures as point clouds and surface meshes to extract quantitative morphological features, including inter-membrane distances, curvature, and surface roughness. It supports both closed and open surfaces, with specific considerations for open surfaces, which are common in cryo-ET due to the missing wedge effect. 

We demonstrate both tools using \emph{in vitro} reconstituted membrane systems containing deformable vesicles with complex geometries, enabling automatic quantitative analysis of MCS and remodeling events such as invagination. While demonstrated here on cryo-ET membrane data, the combined approach is applicable to ROI detection and surface analysis in broader scientific imaging contexts.
\end{abstract}
\medskip



\vspace{0.6em}
]
\makeatother

\section{Introduction}

Cryo-electron tomography (cryo-ET) has emerged as a powerful technique for visualising the three-dimensional (3D) architecture of biological specimens, including \emph{in vitro} reconstituted proteins assemblies, viruses, purified organelles, and cellular structures \emph{in situ}. By preserving samples in a near-native, hydrated state, cryo-ET provides detailed insights into the spatial organisation and interactions of macromolecular complexes and membranes at sub-nanometre resolution \citep{Gemmer2023,Hutchings2018,Nogales2024,Zhang2023}. Of particular interest are organelle membranes, which host distinct protein compositions that support essential cellular functions and undergo dynamic remodelling. The dense and structurally complex cellular context of tomograms, combined with low signal-to-noise ratios and reconstruction artifacts such as the missing wedge, makes reliable membrane segmentation and subsequent quantitative analysis particularly challenging, prompting the development of dedicated computational tools \citep{Grotjahn2025}. Among membrane segmentation tools, TomosegmemTV adopts a classical tensor-voting based approach \citep{MartinezSanchez2014}, while some others provide pre-trained U-nets to be readily deployed or fine-tuned by the user: DeePiCT \citep{deTeresaTrueba2023}, Membrain-seg in Membrainv2 \citep{Lamm2025}, Tardis \citep{Kiewisz2025}, and the commercial software Dragonfly \citep{Makovetsky2018}. Following membrane segmentation, morphological analysis can be performed to extract quantitative features, such as inter-membrane distances and membrane curvature. Existing tools include Pycurv \citep{Salfer2020} and Surface Morphometrics \citep{Barad2023}, which offer options for quantitative membrane morphological analysis. In contrast, Membrainv2 \citep{Lamm2025} focuses on membrane protein analysis by combining membrane segmentation with protein localisation statistics.\par

Despite these advances, segmentation approaches across scientific imaging primarily focus on complete structural entities rather than spatially restricted regions of interest (ROIs). However, many biological and physical questions concern localised domains within continuous and geometrically complex structures that cannot be defined as independent objects and often lack a clear mathematical definition. In current practice, such regions are typically identified indirectly following global segmentation and manual inspection, relying on \emph{post hoc} filtering or manual curation rather than direct detection. Consequently, automated, shape-agnostic direct ROI segmentation has not been established as a general analytical strategy, particularly for domains that require learned representations rather than predefined geometric criteria. In addition, robust automated processing of open surfaces, which are frequently encountered in cryo-ET data due to the missing wedge, remains limited across imaging domains. As cryo-ET datasets increase in size and complexity, scalable tools that enable direct ROI identification and quantitative surface analysis while minimising manual curation are urgently needed.\par

Here, we present TomoROIS and SurfORA, two tools introducing novel direct, shape-agnostic ROI segmentation and automated quantitative surface morphometry. TomoROIS implements a deep learning-based method capable of segmenting both complete structures and explicitly defined ROIs, including regions without predefined borders, enabling targeted analysis based on geometry or spatial context. SurfORA provides automated surface morphometry with dedicated algorithmic support for both closed and, in particular, open surfaces. These tools can be applied independently or sequentially, enabling ROI-guided membrane segmentation followed by automated surface quantification.\par

We demonstrate their capabilities using two \emph{in vitro} reconstituted membrane systems: ROIs between apposed membranes mimicking membrane contact sites (MCS) between organelles in cells \citep{Voeltz2024,Wozny2023}, and ROIs exhibiting membrane deformations of lipid vesicles as observed in intracellular trafficking \citep{Bertin2020,McMahon2005}.\par

\FloatBarrier
\section{Experimental data}

We analysed two \emph{in vitro} cryo-ET datasets. The first consists of reconstituted MCS formed by VAP-A and OSBP(1–407) complexes tethering apposed membranes of lipid nanotubes and vesicles. This dataset was previously reported \citep{deLaMora2021}. On average, each tomogram contains \textasciitilde{}6 membrane contact regions spanning 100–250 nm. Protein densities are visible at vesicle–tube interfaces, although individual tethering proteins (120 kDa) cannot be resolved due to their small size and flexibility. The dataset comprises 58 tilt series of 41 images each, acquired at 3° increments on a Titan Krios (Thermo Fisher Scientific, USA) equipped with a Gatan K2 camera, at a pixel size of 1.7 Å.\par

The second dataset consists of pure lipid vesicles exhibiting membrane invaginations induced by osmotic pressure following a 10-minute incubation at 80\% humidity in the chamber of an EMGP2 Leica plunger. Thirty tilt series were acquired on a Glacios microscope equipped with a Falcon IV camera at a pixel size of 1.55 Å. Tilt series were processed with TomoCHAMPS, our custom automated software integrating drift correction with MotionCor2 \citep{Zheng2017}, Contrast Transfer Function (CTF) estimation with CTFplotter in IMOD \citep{Mastronarde2024}, tilt series alignment using Dynamo \citep{Coray2024}, tomogram reconstruction by weighted back projection in IMOD, and denoising and missing wedge compensation with CryoCARE \citep{Buchholz2018} and Isonet \citep{Liu2022}.\par

\section{Results}

\begin{figure*}[!t]
\centering
\includegraphics[width=0.8\textwidth]{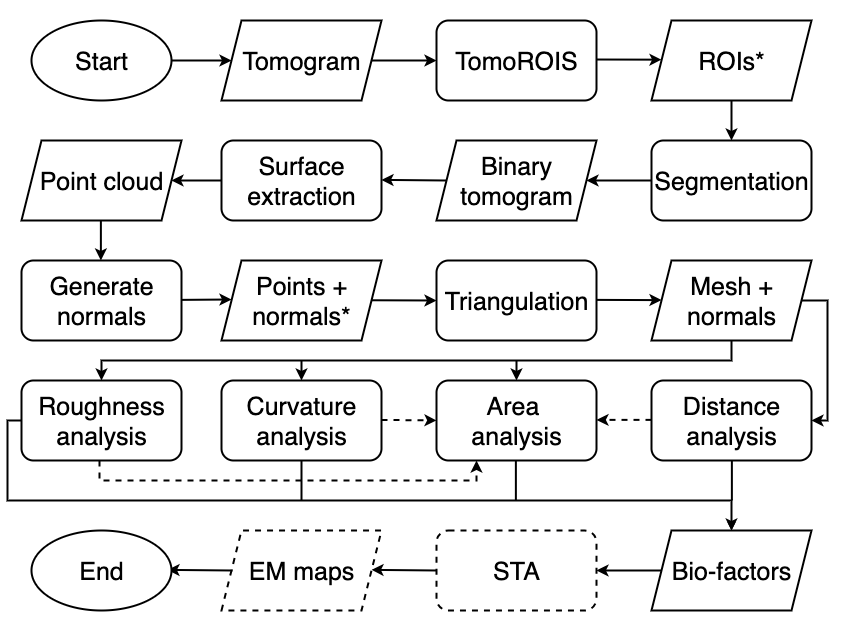}
\caption{Overview of the TomoROIS-SurfORA workflow. An asterisk marks steps with optional GUI based curation, and dotted lines indicate optional workflow steps.}
\label{fig:workflow}
\end{figure*}

We developed Tomography-based Region of Interest Segmentation (TomoROIS) and Surface Topography and Geometrical Analysis (SurfORA) as dedicated software toolkits. Both are designed to allow users either to execute individual functions independently or to follow a recommended workflow sequence, depending on their analytical needs.\par

Figure~\ref{fig:workflow} outlines the TomoROIS–SurfORA workflow followed in this study. After denoising and missing-wedge compensation of tomograms, for example, using CryoCARE \citep{Buchholz2018} and IsoNet \citep{Liu2022}, which are widely adopted in the cryo-ET community. Processing begins with TomoROIS, where users manually annotate ROIs to generate training data for a provided neural network, which can be trained from scratch and subsequently applied for automated ROI segmentation. Comprehensive annotation, visualisation and editing are supported through our Napari-based graphical user interface (GUI) \citep{Chiu2022}. For example, dedicated functions facilitate curation of predicted segmentations by enabling tomogram visualisation with mask overlays and interactive adjustment of threshold parameters.\par

After ROI definition, the region can either be segmented directly within the ROI or used to extract the corresponding subregion from a pre-computed global segmentation. Depending on the structural characteristics of the target, such as membranes, tubular structures, surfaces, or volumetric objects with defined geometry, an appropriate external segmentation tool may be applied.\par

The ROI-restricted segmentation is then converted into a surface point cloud within SurfORA, from which consistently oriented normals are computed next. Local orientation inconsistencies, if present, can be manually corrected through our interactive GUI build on the Python library PyVista \citep{Sullivan2019}. The oriented point cloud can then be meshed and used for quantitative surface analysis, including curvature estimation, inter membrane or surface to surface distance measurements, roughness quantification and surface area calculation. These outputs provide structured descriptors of surface topography and geometry that enable comparison between conditions, identification of localised architectural features and inference of underlying physical or biological properties. In addition, these parameters can be exported as filtering or classification features to support subtomogram averaging (STA) by introducing contextual geometric information that may assist in reducing particle heterogeneity. An overview of SurfORA algorithm design and its functionalities demonstrated in this study is provided in supplementary~\hyperref[sec:S3]{S3}.\par

We utilise a mixed scale dense convolutional neural network \citep{PeltSethian2018} (MSDCN) from random initialisation to perform automated ROI segmentation within TomoROIS. MSDCN consists of densely connected convolutional layers operating across multiple effective receptive field sizes without the use of explicit pooling layers or skip connections. This configuration enables it to integrate both local membrane features and their surrounding contextual environment while remaining compact and trainable on the limited volumes of annotated cryo-ET data typically available.\par

In this context, annotation quality refers to how well examples capture the intended spatial context rather than pixel precise delineation of anatomical boundaries. TomoROIS is not designed to recover sharply defined borders but to identify contextual regions with user-defined margins relative to surrounding structures, where boundaries are inherently flexible. Accordingly, the task addressed by TomoROIS lies at the interface between object detection, which identifies ROI, and segmentation, which generates masks with appropriate spatial extent.\par

Two complementary metrics were used to evaluate TomoROIS performance. To assess the object detection component, true positives, false positives and false negatives were quantified at the level of ROI identification. To evaluate segmentation accuracy in datasets where shape is relevant, Dice and intersection over union scores were computed by comparing predicted masks with manually corrected references. Binary cross entropy was used as the training loss function.\par

\begin{figure*}[!t]
\centering
\includegraphics[width=0.6\textwidth]{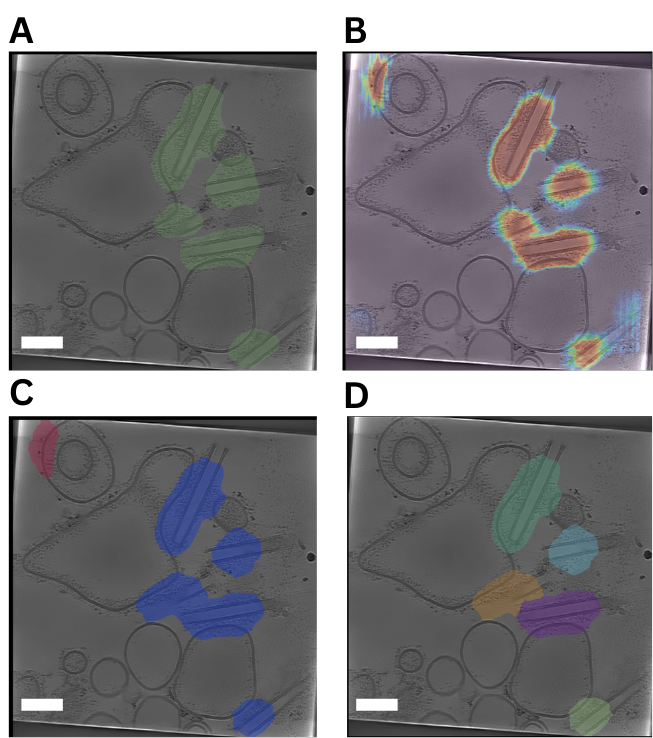}
\caption{Region of interest segmentation with TomoROIS for MCS. A: Manually annotated ROIs of two interacting membranes of vesicles and tubes bridged with protein complexes. B: Predicted ROIs generated by MSDCN are shown on the same tomogram slice. Pixel-wise prediction confidence is visualised using a colour scale, where red corresponds to the highest confidence, gradually shifting through yellow and green to blue with decreasing confidence. C: A confidence threshold was applied to retain only pixels predicted with high confidence as ROIs. Correct predictions are highlighted in blue, while false predictions are shown in red, including a ‘vesicle-in-vesicle’ region that was incorrectly classified as a MCS. D: The retained blue regions were first separated into non-connected components and then further divided using a watershed-based splitting step implemented in TomoROIS to separate touching ROIs. Each resulting ROI is displayed in a distinct colour. Scale bar: 50nm.}
\label{fig:2}
\end{figure*}

We applied TomoROIS followed by SurfORA to two datasets containing geometrically distinct target subregions. In the first dataset, the targets correspond to interaction regions between two separate objects, namely vesicles and lipid nanotubes in \emph{in vitro} reconstituted MCS mimicking ER–Golgi contacts (Fig.~\ref{fig:2}). In the second dataset, the targets consist of local membrane deformations and invaginations within individual lipid vesicles, resembling protein induced membrane remodelling (Fig.~\ref{fig:5}). The MCS dataset was acquired on a 300 kV cryo electron microscope equipped with an energy filter, whereas the vesicle invagination dataset was collected at 200 kV without an energy filter and exhibits a lower signal to noise ratio.\par

\subsection{Region of interest segmentation}

The MSDCN was initially trained on 10 annotated tomograms from the MCS dataset, comprising approximately 50 interaction ROIs between vesicles and tubes. The trained model was then applied to an additional 10 tilt series, which were curated and used for subsequent fine-tuning before deployment on the full dataset of 58 tilt series. Representative results are shown in Fig.~\ref{fig:2}A–E. In Fig.~\ref{fig:2}A, the manually annotated ground truth highlights the membrane region where a vesicle approaches a lipid nanotube, shown in bright green on a denoised and missing wedge compensated tomographic slice. Fig.~\ref{fig:2}B shows the MSDCN prediction for the same region, with prediction confidence visualised as a colour gradient ranging from red for high-confidence predictions to blue for low-confidence predictions. This tomogram was held out from training and validation and used solely for testing. To retain reliable predictions, a confidence threshold was applied as shown in Fig.~\ref{fig:2}C, followed by automatic post-processing including denoising, smoothing and morphing. High confidence regions are shown in blue, whereas erroneous predictions are shown in red. Following thresholding, predicted ROIs were separated into disconnected components and, when required, further subdivided using a watershed procedure implemented in TomoROIS, as illustrated in Fig.~\ref{fig:2}D, where distinct ROIs are displayed in different colours.\par

Inspection at this stage indicated that segmentation performance was adequate for downstream analysis, with only minimal manual correction required to remove occasional false positives. Two characteristic error patterns were observed. First, the network occasionally misclassified isolated nanotubes as MCS. Second, in the invagination dataset, MSDCN sometimes failed to detect invaginations when the deformation was oriented predominantly in the XY plane rather than the more frequently observed XZ plane (see below Fig.~\ref{fig:5}). To mitigate such errors, TomoROIS allows an optional training penalty that down-weights specific mispredicted patterns, reducing model confidence in ambiguous configurations. When enabled, this penalty decreased the confidence of recurrent false positives, facilitating their filtering while preserving high-confidence true positives. This improved the interpretability and robustness of the final segmentation.\par

In the MCS dataset, TomoROIS achieved a false positive rate of 17\% among predicted regions and a false negative rate of 3\% across all 388 ROIs. In practice, removing false positives was faster than manually adding missed regions. Shape agreement was high, with Dice and intersection over union scores of 0.89 and 0.83, respectively.\par

In the invagination dataset, where predicted shapes were consistent with the homogeneous spherical ROIs, the false positive rate was 10.5\% and the false negative rate 1\% across 937 ROIs.\par

Overall, these results indicate that MSDCN provides a practical and adaptable framework for ROI segmentation within TomoROIS. The model can be trained from scratch with limited annotated data and iteratively refined through curation and fine tuning. Rather than training once on a fixed dataset, users may progressively curate additional predictions and incorporate them for further fine-tuning until satisfactory performance is achieved.\par

\subsection{Membrane segmentation and surface extraction}

\begin{figure*}[!t]
\centering
\includegraphics[width=\textwidth]{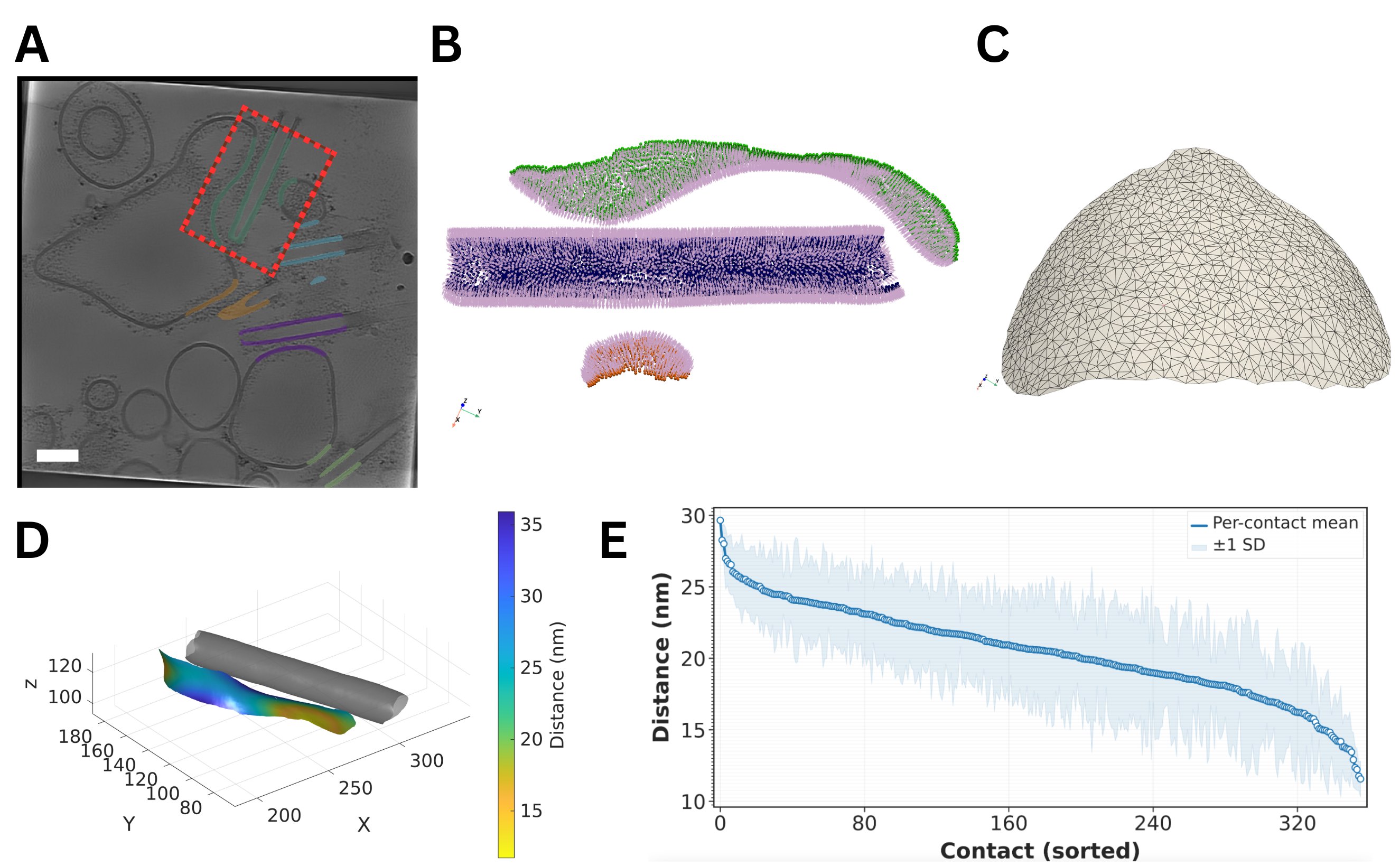}
\caption{Membrane segmentation followed by surface analysis with SurfORA. A: Membrane segmentation extracted using refined ROIs shown in Fig. 2D. Segmented membranes are colour-coded by ROI identity, with the red square indicating the example region used to demonstrate SurfORA utilities. B: Surface point clouds corresponding to contacting regions of two lipid vesicles and one tube displayed as single-layer dense point sets. Non-connected components are coloured green, blue, and orange. Consistently oriented surface normals are shown as pink vectors, with vesicle normals pointing toward the tube and tube normals pointing toward the vesicles. C: Mesh generated from the smaller vesicle membrane. D Inter-membrane distance mapping between the vesicle (colour-mapped from yellow to blue by separation distance) and tube (grey) meshed surfaces. E: Distribution of averaged inter-membrane distances across >350 contact regions from 50 tomograms. The average distance is shown as a dark blue dot, with $\pm$1 standard deviation bands in light blue. Scale bar: 50 nm.}
\label{fig:4}
\end{figure*}

To extract membranes from a representative tomogram containing MCS, the refined ROIs shown in Fig.~\ref{fig:2}D were used as spatial constraints. Global membrane segmentation was performed using the latest pretrained MemBrain-seg U-Net model optimised for denoised cryo-ET data \citep{Lamm2025}. A high confidence threshold was applied to generate a binary membrane mask, which was then restricted to the TomoROIS-predicted ROIs to isolate membranes involved in MCS (Fig.~\ref{fig:4}A). The red square indicates the region used to illustrate key steps of the SurfORA workflow, including surface extraction, normal estimation and orientation, meshing, and inter-membrane distance mapping. Prior to extraction, the ROI-restricted binary mask is denoised to remove salt and pepper noise, small artefactual components, and protrusive segmentation artefacts using binary opening and closing operations on the voxel grid.\par

SurfORA provides two complementary surface extraction strategies. The medial surface is computed by converting the segmented volume into a point cloud and locally projecting it onto a single-layer surface using moving least squares (MLS) projection, combining principal component analysis with quadratic surface approximation. The resulting surface is refined through adaptive Poisson-disc sampling, with curvature-adaptive densification and homogenization generating a dense, curvature-aware, and averaged representation of membrane geometry (described in supplementary~\hyperref[sec:medial_surface_extraction]{S3a}).\par

In contrast, the isosurface is reconstructed from the binary volume by computing a signed distance field (SDF) via the Euclidean distance transform, followed by mean-curvature flow smoothing with gap preservation. A mesh is extracted using the Marching Cubes algorithm and refined through geodesic normal smoothing and adaptive Poisson-disc resampling. Point density is increased in high-curvature regions using curvature-adaptive sampling, and normals are consistently oriented based on the SDF. This approach preserves inner and outer membrane boundaries and ensures accurate capture of local membrane geometry compared to an averaged medial surface representation (described in supplementary~\hyperref[sec:isosurface_extraction]{S3b}).\par

For MCS, the medial surface representation was sufficient, as the interacting membranes are predominantly planar and exhibit limited curvature. For invaginated vesicles, isosurface extraction was employed to resolve pronounced curvature and to distinguish between the inner and outer membrane surfaces.\par

\subsection{Surface-normal estimation and consistent orientation}

SurfORA computes consistent normal orientations for membrane surface point clouds using either dipole normal propagation, which performs reliably for relatively simple geometries such as those in the MCS dataset as shown in Fig.~\ref{fig:4}B \citep{Metzer2021}, or our dedicated geodesically informed heat-based \citep{Sharp2018} propagation scheme that provides robust orientation across geometrically and topologically complex point clouds, as of invaginated vesicles.\par

\begin{figure*}[!t]
\centering
\includegraphics[width=0.7\textwidth]{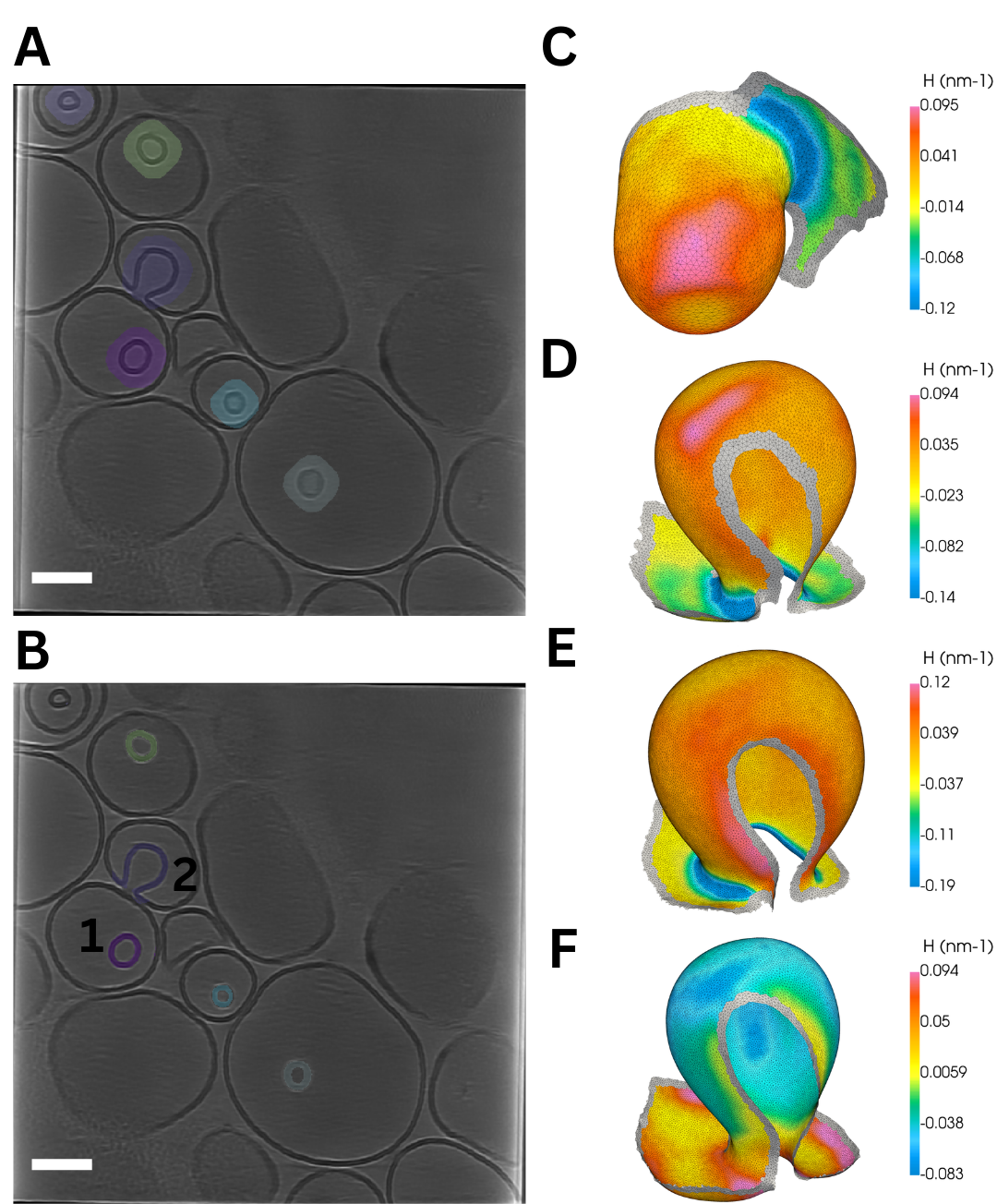}
\caption{Direct detection of membrane invagination with TomoROIS and surface analysis with SurfORA. A: Membrane invagination predicted by TomoROIS. Invaginating vesicles with openings aligned along either the Z or X/Y axes are shown. Each invagination event is uniquely colour-coded. B: Membrane segmentation within ROIs containing invaginating membranes. C, D: Mean curvature maps for medial surfaces of two example invaginating membranes numbered 1, 2 in B. Outer (E) and inner (F) isosurface of invagination numbered 2 with their respective curvature mapping. The colour scale for positive and negative curvature is provided for each mapping plot. Scale bar: 50 nm.}
\label{fig:5}
\end{figure*}

This approach combines local normal estimation with global propagation over a geodesically weighted neighbourhood graph to enforce orientation consistency across branching, irregular, or topologically complex surfaces (described in supplementary~\hyperref[sec:normal_orientation]{S3c}). It remains robust even for unorganised point clouds in which coordinates deviate from an ideal surface. The fully CPU-parallelised implementation remains computationally efficient for dense and complex point clouds and, in the cases examined here, enabled stable normal orientation without any manual correction, thereby supporting reliable mesh construction and curvature mapping in the invaginated vesicles shown in Fig.~\ref{fig:5}C-D.\par

As shown in Fig.~\ref{fig:4}B, the resulting vesicle normals consistently point toward the tube, while tube normals point toward the vesicles, establishing correct relative orientation for downstream geometric analysis. Occasional local inconsistencies can be interactively corrected through SurfORA’s dedicated GUI, which provides comprehensive 3D visualisation capabilities (supplementary Fig.~\ref{fig:S1}). The interface supports interactive 3D point cloud editing using mouse-based and polygon-based selection tools for flexible individual or collective point deletion, normal flipping, and point cloud class labeling.\par

\subsection{Gap-preserving meshing and isosurface partition}

Meshes are generated from medial and isosurface point clouds with consistently oriented normals using ball pivot surface reconstruction with automatic radius scanning to preserve geometric fidelity. To prevent spurious bridging across gaps, gap-aware filtering based on the Separating Axis Theorem can be applied directly using the point cloud with a small tolerance margin given sufficient homogeneous density, removing unsupported false triangles (described in supplementary~\hyperref[sec:ball_pivot]{S3d}). In the example ROI, the reconstructed vesicle mesh (Fig.~\ref{fig:4}C) exhibits nearly uniform vertex spacing. This method was applied to over 700 membrane point clouds from 50 tomograms with negligible artefacts and no manual correction required.\par

For open surfaces, the inner and outer isosurfaces are separated by first reconstructing a medial surface mesh using ball pivoting. A Poisson-reconstructed mesh is then generated from the medial surface point cloud and corresponding normals, which are extracted from the ball-pivot reconstructed mesh. This Poisson mesh fills internal holes and extends open boundaries (described in supplementary~\hyperref[sec:poisson_proxy]{S3e}). It then acts as a proxy, intersecting with the isosurface mesh (reconstructed using ball-pivoting) to partition it into inner and outer components (described in supplementary~\hyperref[sec:split_isosurface]{S3f}). The ball-pivot reconstructed medial surface mesh along with the split inner and outer isosurfaces are visualised in supplementary Fig.~\ref{fig:S2}, overlaid on the membrane segmentation and tomogram. For closed surfaces, inner and outer isosurfaces are intrinsically separated and require no additional processing. Mesh smoothing can be optionally applied, for open medial surfaces or split isosurfaces, mild edge-preserving smoothing (damped Laplacian smoothing) is used, which fixes boundary vertices and applies low-amplitude smoothing to the interior to reduce noise without causing shrinkage or geometric distortion.\par

\subsection{Inter-membrane distance quantification}

Inter-membrane distances were computed on the vesicle vertex against the tube mesh (described in supplementary~\hyperref[sec:distances]{S3g}). The vesicle surface was colour-mapped from low (yellow) to high (blue) separations (Fig.~\ref{fig:4}D); the tube surface is shown in grey. Across more than 350 contact regions from 50 tomograms, separations ranged from 10 to 30 nm, with most distances between 15 and 25 nm. These values are consistent with those previously reported from manually segmented tilt series \citep{deLaMora2021}. For each contact, the mean separation is plotted with a $\pm$ one standard deviation band in Fig.~\ref{fig:4}E, with contacts ranked by decreasing mean distance.\par

\subsection{Morphological analysis of invaginated membranes with SurfORA}

Membrane invagination is a common and biologically relevant event, yet its mathematical description is challenging due to the geometric variability it exhibits. Such structures cannot be reliably detected by applying simple curvature thresholds to globally segmented membranes, particularly in biologically complex systems where baseline morphology is already heterogeneous.\par

To enable automated detection, approximately 50 invagination events from five denoised and missing-wedge compensated tomograms were manually annotated as ROIs and used to train TomoROIS. The trained MSDCN-based model was applied to additional tomograms, iteratively curated and fine-tuned, and subsequently deployed on the full dataset of 30 tomograms of invaginating vesicles. Invaginations varied in orientation, with openings aligned along either the Z or X/Y axes, and ranged from 20 to 80 nm in diameter (Fig.~\ref{fig:5}A). Membrane segmentation was subsequently performed using Membrain-seg \citep{Lamm2025} within these predicted ROIs, yielding membrane segmentation results as shown in Fig.~\ref{fig:5}B.\par

The segmented invaginations were processed with SurfORA following the same workflow as for the MCS dataset, including surface extraction, normal orientation, and meshing. Unlike the predominantly planar MCS membranes, invaginated vesicles required isosurface extraction to accurately represent their pronounced curvature. Curvature analysis was first performed on the medial surface to capture the averaged membrane geometry (Fig.~\ref{fig:5}C, D). To distinguish differences between the two membrane sides, curvature was subsequently analysed on the corresponding inner and outer isosurfaces of the invagination shown in Fig.~\ref{fig:5}D, which more faithfully represent the boundary of each leaflet (Fig.~\ref{fig:5}E, F).\par

Signed Gaussian and mean curvature were estimated for each triangulated membrane mesh by fitting a quadratic surface (Monge form) in a local tangent frame aligned with the input vertex normal. For each vertex, a weighted geodesic neighbourhood was defined using radius-based selection in physical units (e.g. nm), with geodesic distances computed via the heat method \citep{Sharp2018} to enable efficient parallelised computation on dense meshes. Multi-scale radii were evaluated to capture curvature across scales. The optimal radius was selected using combined relative and absolute stability criteria, ensuring that curvature estimates varied minimally between consecutive scales while remaining within an absolute tolerance threshold. This stability-based selection reduced sensitivity to noise and scale dependent artefacts. Boundary vertices were excluded due to their susceptibility to instability (described in supplementary~\hyperref[sec:curvature]{S3g}). Here, positive mean curvature values correspond to outward bending consistent with vertex normal orientation. Per vertex or per face curvature maps can be respectively computed for three-dimensional visualisation and further inspection.\par

\section{Discussion}

Most segmentation pipelines remain object centric. In cryo-ET, typical targets are complete organelles such as mitochondria, endosomes or lysosomes, or cytoskeletal elements such as microtubules, which are segmented globally and analysed afterwards. Within this framework, ROIs are generally derived indirectly, for example, through distance thresholds, curvature filtering or manual selection following full object segmentation. The ROI is therefore treated as a secondary construct rather than as a primary segmentation objective.\par

TomoROIS establishes direct, shape-agnostic ROI segmentation as a primary task. Instead of assuming that biologically relevant units correspond to geometrically closed objects, it enables ROIs to be defined through spatial context, including configurations involving multiple interacting structures or domains with flexible margins that lack clear geometric boundaries. By learning contextual patterns from limited annotations, ROI definition becomes hypothesis-driven and adaptable, rather than constrained by predefined object classes.\par

MSDCN \citep{PeltSethian2018,SegevZarko2022} was selected for TomoROIS due to its lightweight and efficient architecture, requiring primarily the number of layers as a meaningful user-defined design parameter. This reduced architectural complexity enables training from scratch on limited annotated cryo-ET data while supporting iterative retraining for newly defined ROI types. Through dilated convolutions, MSDCN integrates local membrane features with their surrounding context, allowing multi-scale feature aggregation without increasing model complexity—an important advantage for ROI detection in cryo-ET, where membrane boundaries are often ambiguous and pixel-perfect annotations are not always required. Compared to architectures such as U-Net \citep{Ronneberger2015}, which involve additional architectural and training design choices, MSDCN reduces the number of tunable components and limits sensitivity to user-defined settings. 

A remaining challenge arises in MCS datasets with extensively overlapping ROIs. Although watershed-based approaches can separate blob-like touching structures, they do not recover shared regions and struggle with elongated geometries characteristic of MCS, making full separation of partially fused interaction sites into distinct events difficult. Future improvements will therefore focus on more robust instance-level separation strategies tailored to overlapping ROIs.

SurfORA provides automated surface extraction and quantitative morphometry from segmented volumetric data by generating medial or isosurface point-cloud representations that can be reconstructed into meshes depending on the analytical objective. It supports both closed and open surfaces, branched geometries, and unorganised point clouds, with surface reconstruction, normal orientation, meshing, and geometric measurements largely automated and parallelised on CPU for accessible and scalable processing. 

Across biologically relevant membrane geometries, SurfORA achieved consistent normal orientation and robust meshing. Minor ambiguities were observed only in highly non-manifold or artificially constructed geometries, such as intersecting planes or Möbius-like surfaces, where few local inconsistencies in normal direction may arise near intersection or topological transition regions. In irregular point clouds where points deviate from an ideal surface and input normals are poorly estimated, SurfORA was nevertheless able to enforce globally consistent inwards or outwards orientation and apply normal smoothing to improve coherence with the expected surface geometry. Further systematic evaluation on increasingly irregular and topologically complex datasets will help define its operational limits and expand its applicability.

The applicability of TomoROIS and SurfORA is demonstrated across MCS and membrane invaginations, capturing both interaction interfaces and locally remodelled geometries. In MCS datasets, ROI detection enabled targeted inter-membrane distance analysis, whereas in invaginations SurfORA resolved distinct positive and negative curvature domains corresponding to neck and bud regions. The ability to extract medial or isosurface representations further allowed geometric analysis at different levels of structural detail, including separate assessment of inner and outer membrane surfaces when required. Both tools provide dedicated GUIs for visualisation and curation, supporting user-guided validation alongside automated processing.\par

Beyond these examples, the framework is not restricted to membranes and can extend to other structural components such as cytoskeletal elements, condensates, or viral particles in cryo-ET. MSDCN has demonstrated applicability in \emph{in situ} cryo-ET, including segmentation of microtubules, apical vesicles, rhoptries, and micronemes in \emph{Toxoplasma gondii} \citep{SegevZarko2022}, supporting broader use with limited annotated data when sufficient signal-to-noise ratio is achieved. Incorporating geometric descriptors such as membrane curvature as quantitative “bio-factors” may further reduce particle heterogeneity in subtomogram averaging workflows. More broadly, the underlying principles are applicable to ROI detection and surface-based analysis in other volumetric imaging modalities beyond cryo-ET.\par

\section*{Acknowledgments}

This work was supported by Centre National de La Recherche Scientifique, Institut Curie, Agence Nationale de La Recherche (ANR-21-CE13-0021-01) (to M.D D.L.) and by a PhD fellowship from the PSL University and EURECA cofound European program (XC).  We thank the Cell and Tissue Imaging core facility (PICT IBiSA), Institut Curie, member of the French National Research, Infrastructure France-BioImaging (ANR-24-INBS-0005 FBI BIOGEN, \url{https://ror.org/01y7vt929}) and Region Ile de France (Sesame 2018 3D EM/CLEM EXO039200). We thank Ilyes Hamitouche, Daniel Castaño-Díez and Charles Kervrann for fruitful discussions.\par

\section*{Author contribution}

Xingyi Cheng: Conceptualization, Methodology, Investigation, Writing -- Original Draft, Writing -- Review \& Editing.
Julien Maufront: Conceptualization, Methodology, Investigation, Writing -- Review \& Editing.
Aurélie Di Cicco: Investigation.
Daniël M. Pelt: Investigation, Writing -- Review \& Editing.
Manuela Dezi: Funding acquisition, Supervision.
Daniel Lévy: Conceptualization, Funding acquisition, Supervision, Writing -- Original Draft, Writing -- Review \& Editing.

\section*{Declaration of competing interest}

The authors declare that they have no competing financial interests or personal relationships that could have influenced the work reported in this paper.\par

\section*{Data and code availability}
Detailed algorithmic implementation, source code, and tutorial documentation will be made publicly available upon journal acceptance. In the interim, materials can be provided upon reasonable request.\par

\bibliographystyle{unsrtnat}
\setlength{\bibsep}{2pt}

\setcounter{figure}{0}
\renewcommand{\thefigure}{S\arabic{figure}}

\begin{figure*}[!t]
\centering
\includegraphics[width=\textwidth]{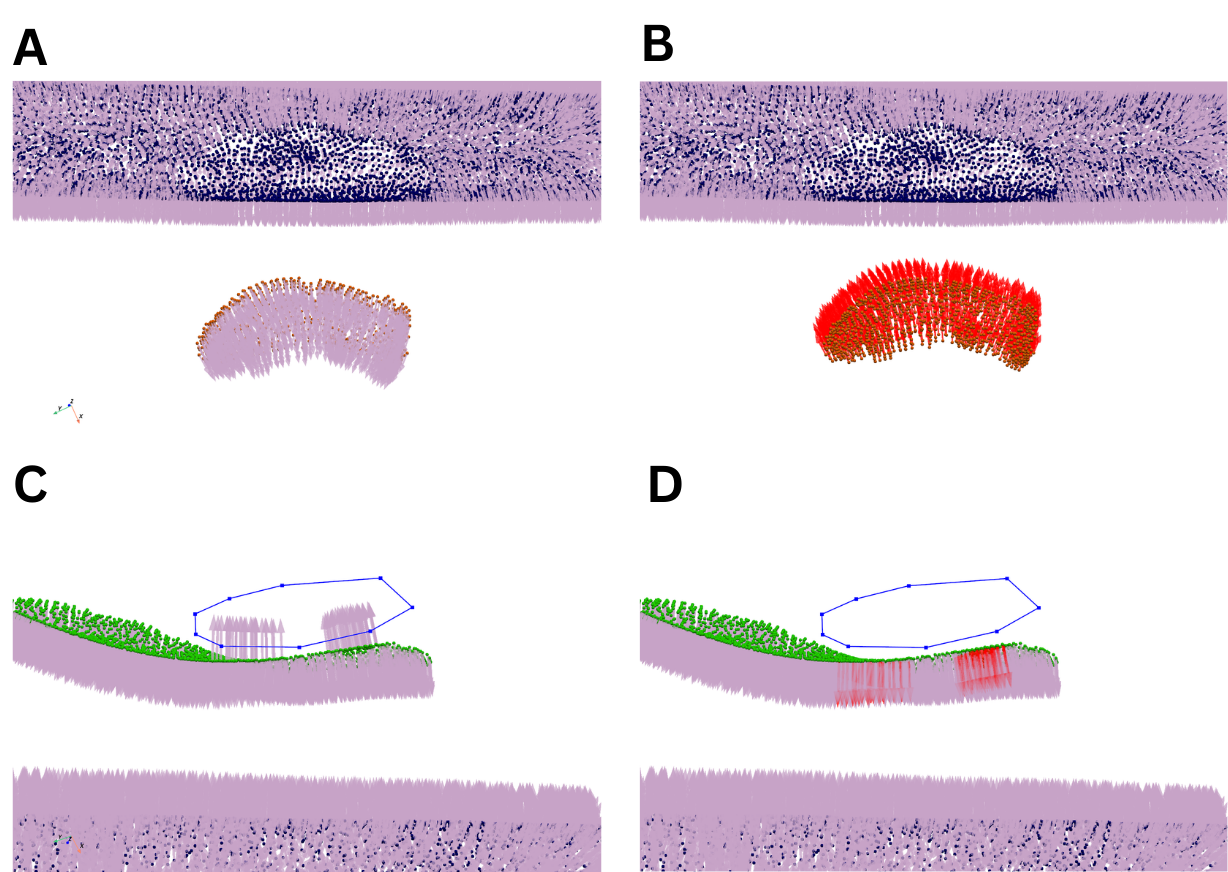}
\caption{Interactive visualisation and curation of point cloud with normals. Manual curation of membrane point clouds and their associated normals in interactive GUI. Case 1: inversion of normals of a labeled object. A: non-corrected normals of the tube (label 1, blue points) and the vesicle (label 2, orange points) are in pink.  B: normals of label 2 after flipping shown in red. Case 2: local curation of normals. C: Selection of normals displaying wrong orientation within a manually drawn polygon in dark blue. D: Selected normals after flipping shown in red.}
\label{fig:S1}
\end{figure*}
\FloatBarrier

\begin{figure*}[!t]
\centering
\includegraphics[width=\textwidth]{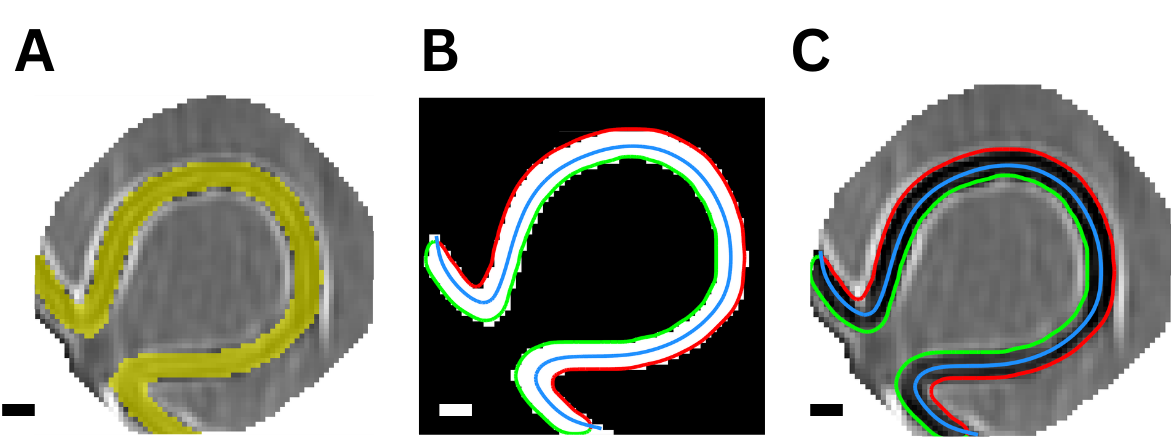}
\caption{Medial and isosurface extraction of an invaginated vesicle. Membrane segmentation and corresponding surface representations for a representative invaginated vesicle. A: Membrane segmentation overlaid on the tomogram. B: Extracted medial surface shown in blue, and split isosurfaces corresponding to the inner (green) and outer (red) membrane boundaries, overlaid on the membrane segmentation. C: The extracted medial surface (blue) and inner (green) and outer (red) isosurfaces overlaid directly on the tomogram. Scale bar: 5 nm.}
\label{fig:S2}
\end{figure*}
\FloatBarrier

\clearpage
\appendix

\begingroup
\setlength{\parskip}{0.6em}
\setlength{\parindent}{0pt}

\section*{S3: SurfORA algorithm briefing}
\label{sec:S3}

\subsection*{Workflow overview}

Given a curated binary membrane segmentation within a region of interest (ROI), SurfORA derives geometric surface models and quantitative measurements via the following stages:

\begin{enumerate}
\item medial surface extraction as a dense single-layer mid-surface point cloud,
\item isosurface extraction as a dense single-layer point cloud representing the inner and outer membrane boundaries,
\item global orientation of locally estimated surface normals,
\item ball-pivot surface reconstruction with gap-aware filtering,
\item optional hole-filled, boundary-extended medial proxy via Poisson surface reconstruction,
\item separation of the isosurface into inner and outer membrane leaflets,
\item inter-surface distance and signed curvature analysis.
\end{enumerate}

\paragraph{Notation}
\begin{itemize}
\item Voxel centres are denoted by $\mathbf{s}_i \in \mathbb{R}^3$; the voxelised membrane support is $\mathcal{S}=\{\mathbf{s}_i\}$.
\item Point clouds are denoted by calligraphic sets (e.g.\ $\mathcal{P}\subset\mathbb{R}^3$).
\item A triangulated surface is denoted by $\mathcal{T}=(\mathcal{V},\mathcal{F})$, with vertex set $\mathcal{V}$ and triangular face set $\mathcal{F}$.
\item Signed distance fields are denoted by $\phi$.
\item Surface normals at vertices are denoted by $\mathbf{n}_i$.
\item Geodesic distances are denoted by $g_i$.
\item All geometric quantities (distances, curvatures, radii) are expressed in physical units according to the supplied voxel size.
\end{itemize}

\section{Medial surface extraction}
\label{sec:medial_surface_extraction}

The objective of this stage is to approximate a geometrically smooth mid-surface from a potentially multi-voxel-thick membrane segmentation.

\paragraph{Input.}
A voxelised membrane support
\[
\mathcal{S}=\{\mathbf{s}_i\}\subset\mathbb{R}^3
\]
defined by occupied voxel centres.

\paragraph{Iterative quadratic projection.}
An initial single-layer approximation is obtained via moving least-squares projection.
For each point, a $k$-nearest neighbourhood is extracted and a local orthonormal principle component analysis (PCA) frame
$(\hat{\mathbf{t}}_1,\hat{\mathbf{t}}_2,\hat{\mathbf{n}})$ is computed.
In this frame, the surface is locally expressed in Monge form
\[
z(x,y)=ax^2+2bxy+cy^2+dx+ey,
\]
with coefficients estimated by weighted least squares.
Points are iteratively reprojected onto the fitted quadratic surface,
yielding a single-layer mid-surface approximation.

\paragraph{Curvature-adaptive densification.}
Principal curvatures $(k_1,k_2)$ are obtained from
\[
\mathrm{Hess}(z)=
\begin{pmatrix}
2a & 2b \\
2b & 2c
\end{pmatrix},
\qquad
k_{\max}=\max(|k_1|,|k_2|).
\]
Local surface patches are generated in tangent coordinates using curvature-adaptive sampling,
with spacing inversely proportional to $k_{\max}$, subject to lower and upper bounds to prevent oversampling in near-planar regions and instability in high-curvature or noisy areas.
Candidate points are retained only if supported by the underlying voxel segmentation,
preventing extrapolation beyond the membrane support.

\paragraph{Single-layer enforcement and homogenisation.}
A curvature-aware thickness constraint is applied by discarding points whose displacement along the estimated local normal exceeds an adaptive threshold, enforcing a single-layer representation.
The resulting set is homogenised using variable-radius Poisson-disc sampling, followed by small-component removal and a final MLS reprojection.
Points lying outside the voxelised membrane support are discarded.

\paragraph{Output.}
A dense, curvature-adaptive medial point cloud
\[
\mathcal{P}\subset\mathbb{R}^3,
\]
approximating the geometric membrane mid-surface at the selected fitting and sampling scales.
\section{Isosurface extraction}
\label{sec:isosurface_extraction}

The objective of this stage is to construct a smooth geometric representation of the membrane boundaries from the voxelised support while preserving thin inter-membrane gaps.

\paragraph{Signed distance field.}
A binary volume $\mathrm{vol}$ is constructed from the voxel support $\mathcal{S}$ on a regular grid with spacing equal to the voxel size.
The signed distance field (SDF) is defined as
\[
\phi = \mathrm{EDT}(\neg \mathrm{vol}) - \mathrm{EDT}(\mathrm{vol}),
\]
where $\mathrm{EDT}$ denotes the Euclidean distance transform.
With this convention, $\phi<0$ inside the segmented membrane support and $\phi>0$ outside.
Optionally, the SDF may be resampled to a finer grid prior to smoothing.

\paragraph{Regularisation with gap preservation.}
To reduce voxelisation artefacts, the SDF is first Gaussian-smoothed and subsequently evolved by mean-curvature flow,
\[
\phi_{t+1}
=
\phi_t
+
\Delta t\,|\nabla\phi_t|\,
\nabla\!\cdot\!\left(\frac{\nabla\phi_t}{|\nabla\phi_t|}\right),
\]
which attenuates high-frequency noise while preserving large-scale geometric structure.

To prevent artificial merging across thin membrane gaps, an obstacle constraint is enforced after each iteration:
\[
\phi \leftarrow \max(\phi,\phi_{\mathrm{ref}}),
\]
where $\phi_{\mathrm{ref}}$ denotes the original SDF (after any resampling).
This guarantees that the evolving level set cannot move inward relative to the original segmentation, thereby preserving background gaps.

\paragraph{Surface extraction and sampling.}
The zero level set $\phi=0$ is extracted via marching cubes, yielding a triangulated surface $\mathcal{T}$.
Mesh vertex normals are computed and smoothed intrinsically to reduce discretisation artefacts.
The surface is then resampled either by uniform Poisson-disc sampling (approximately constant spacing) or, optionally, by curvature-adaptive sampling that increases point density in regions of high bending while enforcing a variable minimum-distance constraint to prevent clustering.

\paragraph{Normal orientation.}
For each sampled point $\mathbf{p}$ with provisional normal $\mathbf{n}$, the signed distance is evaluated at $\mathbf{p}$ and at a small offset along the normal direction.
Normals are flipped if
\[
\phi(\mathbf{p}+\varepsilon\mathbf{n}) < \phi(\mathbf{p}),
\]
ensuring alignment with the outward direction (increasing signed distance) under the adopted sign convention.

\paragraph{Output.}
A dense isosurface point cloud
\[
\mathcal{P}_{\mathrm{iso}} \subset \mathbb{R}^3
\]
representing the inner and outer membrane boundaries, with consistently oriented normals.

\section{Consistent normal orientation}
\label{sec:normal_orientation}

The objective of this stage is to obtain globally sign-consistent surface normals from locally estimated, sign-ambiguous directions.

\paragraph{Local normal estimation.}
Initial unoriented normals $\mathbf{n}_i^{(0)}$ are estimated by local quadratic (JET) surface fitting.
For each point $\mathbf{p}_i$, a $k$-nearest neighbourhood is extracted and expressed in a local PCA frame.
A quadratic height function is fitted in least-squares sense, and the normal is corrected using first-order slope terms to obtain a robust local estimate.
Normals are unit-normalised but remain sign-ambiguous.

\paragraph{Geodesically weighted neighbourhood graph.}
To achieve global sign consistency, a sparse intrinsic neighbourhood graph is constructed.
Geodesic distances $g_i$ from a chosen seed vertex are computed using the heat method on the point cloud.

For each $k$-nearest neighbour pair $(i,j)$, an edge is retained only if
\[
|\langle \mathbf{n}_i^{(0)}, \mathbf{n}_j^{(0)} \rangle|
\ge \tau,
\]
preventing connections across strongly misaligned sheets or branches.
Edge reliability is defined as
\[
w_{ij}
=
\exp\!\left(
-\frac{|g_i-g_j|}
{\alpha\,\|\mathbf{p}_i-\mathbf{p}_j\|}
\right)
|\langle \mathbf{n}_i^{(0)},\mathbf{n}_j^{(0)}\rangle|,
\]
which favours locally consistent and geodesically plausible connections while down-weighting shortcut edges.

\paragraph{Maximum-consistency orientation.}
Normals are oriented by traversing a maximum-consistency spanning tree.
Starting from a seed vertex, edges are processed in decreasing reliability order.
Each visited normal is flipped if necessary to agree with its parent.
Disconnected components are oriented independently when present.

Residual inconsistencies are further reduced by iterative geodesically weighted sign voting over the same graph.

\paragraph{Geodesic normal smoothing.}
After sign resolution, normals are smoothed intrinsically, where $\alpha$ (smoothing parameter) controls the degree of smoothing applied to the normal vector $\mathbf{n}_i$ according to the following equation:
\[
\mathbf{n}_i \leftarrow
\alpha \mathbf{n}_i
+
(1-\alpha)
\frac{\sum_j w_{ij}\mathbf{n}_j}{\sum_j w_{ij}},
\]
and the result is followed by renormalisation to maintain unit length.
Smoothing preserves orientation consistency while reducing high-frequency noise.

\paragraph{Output.}
A set of globally consistent, smoothly varying normals
robust to open surfaces, folds, moderate branching,
and non-uniform sampling.

\section{Ball-pivot meshing with gap-aware filtering}
\label{sec:ball_pivot}

The objective of this stage is to reconstruct a triangulated surface $\mathcal{T}$ that faithfully represents the membrane geometry while preventing artefactual connections across unsupported regions.

Ball-pivoting generates $\mathcal{T}$ from an oriented point cloud using a multi-radius strategy to accommodate variations in sampling density and curvature.

\paragraph{Gap-aware filtering.}
To eliminate spurious bridges, the reconstructed surface is validated against a voxelised support volume.
A binary occupancy volume $\mathrm{vol}$ is constructed either from a provided segmentation mask or, if unavailable, from the original (dense, non-downsampled) point cloud.
Gap voxels are identified by thresholding the Euclidean distance transform of $\mathrm{vol}$ according to a user-defined distance criterion.

For each face, triangle--voxel intersection is evaluated within its axis-aligned bounding box using a Separating Axis Theorem (SAT) test.
Faces intersecting any voxel classified as a gap (and, when a labelled mask is available, voxels of a non-matching label) are discarded.
Isolated faces lacking shared edges are subsequently removed.

\paragraph{Output.}
A triangulated surface geometrically consistent with the underlying membrane support.

\section{Poisson-derived medial proxy}
\label{sec:poisson_proxy}

The objective of this stage is to construct a continuous reference surface used to partition the membrane isosurface into inner and outer components.

Poisson surface reconstruction is applied to the vertices and vertex normals of the ball-pivot mesh $\mathcal{T}$, treated as a point cloud with normals.
Using $\mathcal{T}$ rather than the original medial point cloud provides a surface-based representation with an associated normal field, yielding a stable input for Poisson reconstruction in the presence of local sampling variability in the unstructured point cloud.

The Poisson solve produces a proxy surface $\mathcal{T}_{\mathrm{P}}$ that fills holes and extends open boundaries of $\mathcal{T}$.
To suppress weakly supported regions introduced by the global reconstruction, vertices with very low Poisson density values (e.g., within the lowest quantile, typically around 1\%) are removed together with their incident faces, followed by standard mesh clean-up operations (removal of degenerate elements, duplicated elements, non-manifold edges, and unreferenced vertices).
No additional trimming is applied, as the proxy is intended to remain continuous for subsequent partitioning.

The proxy surface is used exclusively as a geometric separator and does not replace $\mathcal{T}$ for quantitative measurements.

\paragraph{Output.}
A continuous Poisson-derived proxy surface $\mathcal{T}_{\mathrm{P}}$ used solely for isosurface partitioning.
\section{Isosurface separation}
\label{sec:split_isosurface}

The objective of this stage is to partition the membrane isosurface into two separate components representing the inner and outer membrane leaflets.

Let $\mathcal{T}_\mathrm{iso}$ denote the membrane isosurface extracted from the SDF.

\paragraph{Closed membranes.}
If $\mathcal{T}_\mathrm{iso}$ consists of two disconnected components, they are identified directly by connectivity analysis and interpreted as inner and outer membrane leaflets.

\paragraph{Open membranes.}
If $\mathcal{T}_\mathrm{iso}$ forms a single connected component, separation is performed geometrically using the Poisson-derived proxy surface $\mathcal{T}_{\mathrm{P}}$.

The intersection curve between $\mathcal{T}_\mathrm{iso}$ and $\mathcal{T}_{\mathrm{P}}$ is computed by finding the curve where the two surfaces intersect. The isosurface is then split along this curve, resulting in two disjoint sub-meshes corresponding to the inner and outer membrane sides.

The partitioned isosurface $\mathcal{T}_{\mathrm{iso, split}}$ is then decomposed into connected components.

\paragraph{Output.}
Two disjoint triangulated surfaces derived from $\mathcal{T}_\mathrm{iso}$, representing the separated membrane sides.
\section{Distance and Signed Curvature}

The objective of this stage is to compute inter-surface distances (from vertices to the mesh) and signed curvature for the membrane surface meshes, including the medial surface, isosurface, or split isosurfaces.

\subsection*{Inter-surface Distance}
\label{sec:distances}

Given a set of query vertices $\{\mathbf{v}_i\}$ from the vertex set $\mathcal{V}$ of a target triangulated surface $\mathcal{T} = (\mathcal{V}, \mathcal{F})$, the Euclidean distance from each query vertex to the closest point on the surface is defined as the point-to-mesh distance. For each query vertex $\mathbf{v}_i$, this distance is computed as:

\[
d_i = \min_{\mathbf{p} \in \mathcal{T}} \mathrm{dist}(\mathbf{v}_i, \mathbf{p}),
\]
where $\mathrm{dist}(\mathbf{v}_i, \mathbf{p})$ represents the Euclidean distance between the query vertex $\mathbf{v}_i$ and the nearest point $\mathbf{p}$ on the surface $\mathcal{T}$, which could be the nearest point on a triangular face or vertex. These distances are computed in physical units (e.g., nm) and summarised statistically (mean, standard deviation, minimum, and maximum distances) to provide an overview and facilitate comparison across conditions.

\subsection*{Signed Curvature}
\label{sec:curvature}

The goal of this section is to compute the signed curvature at each vertex of the membrane isosurface. Signed curvature provides a measure of how the surface bends locally, with signs determined by the orientation of the surface normals.

\paragraph{Geodesic neighbourhoods.}
Intrinsic neighbourhoods are defined using geodesic distances, computed with the heat method over a discrete set of radii $\{r_s\}_{s=1}^{S}$. These distances define local regions around each vertex, and the radii are expressed in physical units (e.g., nm), consistent with the voxel size and intuitive for interpretation.

\paragraph{Monge fitting.}
In a local frame aligned with the input vertex normal, the surface is expressed as a quadratic function in Monge form:
\[
z(x, y) = ax^2 + 2bxy + cy^2 + dx + ey,
\]
where $(x, y)$ are local coordinates in the tangent plane, and $(a, b, c, d, e)$ are the coefficients determined via weighted least squares. The principal curvatures $k_1$ and $k_2$ are computed from the Hessian matrix of this quadratic surface:
\[
\mathrm{Hess}(z) = \begin{pmatrix} 2a & 2b \\ 2b & 2c \end{pmatrix}.
\]
The mean and Gaussian curvatures are then given by:
\[
H = -\frac{1}{2}(k_1 + k_2), \quad K = k_1 k_2,
\]
with the curvature signs determined by the orientation of the input vertex normal, ensuring consistency with the surface orientation. These curvatures are expressed in inverse physical units (e.g., $\mathrm{nm}^{-1}$, $\mathrm{nm}^{-2}$), based on the input voxel size.

\paragraph{Stability-based scale selection.}
For each vertex, the smallest radius that satisfies the following stability criterion is selected:
\[
|H_s - H_{s-1}| \leq \max\left( \delta_{\mathrm{rel}} \max\left( |H_s|, |H_{s-1}|, \varepsilon \right), \delta_{\mathrm{abs}} \right),
\]
where $H_s$ and $H_{s-1}$ are the mean curvatures at radii $r_s$ and $r_{s-1}$, respectively. The parameter $\varepsilon$ is a small value used to prevent numerical instability, particularly when the curvatures are close to zero. $\delta_{\mathrm{rel}}$ is the relative tolerance, which controls the sensitivity of the curvature difference relative to the magnitudes of the curvatures. $\delta_{\mathrm{abs}}$ is the absolute tolerance, which sets a fixed threshold on the maximum allowable difference between curvatures, independent of their magnitudes. If no radius satisfies this condition, the largest radius with a finite estimate is selected. This choice serves as a conservative measure to reduce the likelihood of instability or overfitting that might occur with smaller radii. Gaussian curvature is reported at the selected radius.

\paragraph{Output.}
A per-vertex confidence score is computed based on the number of neighboring vertices used for curvature fitting and the residual of the quadratic fit. The output consists of the following per-vertex quantities:
\[
(H,\, K,\, r_{\mathrm{used}},\, \mathrm{confidence}),
\]
where $H$ is the mean curvature, $K$ is the Gaussian curvature, $r_{\mathrm{used}}$ is the selected radius for curvature estimation, and $\mathrm{confidence}$ is the reliability score. All quantities are expressed in physical units.

\endgroup

\end{document}